\documentclass[conference]{IEEEtran}
\IEEEoverridecommandlockouts

\usepackage{cite}
\usepackage{amsmath,amssymb,amsfonts}
\usepackage{bm}
\usepackage{algorithmic}
\usepackage{graphicx}
\usepackage{textcomp}
\usepackage{color}
\usepackage{xcolor}
\usepackage{colortbl}
\usepackage{xcolor}
\usepackage{booktabs} 
\usepackage{multirow} 
\def\BibTeX{{\rm B\kern-.05em{\sc i\kern-.025em b}\kern-.08em
    T\kern-.1667em\lower.7ex\hbox{E}\kern-.125emX}}
\begin{document}

\title{Spatiotemporal Causal Decoupling Model for Air Quality Forecasting}

\author{\IEEEauthorblockN{Jiaming Ma\textsuperscript{\dag},
    Guanjun Wang\textsuperscript{\dag},
    Sheng Huang\textsuperscript{\dag},
    Kuo Yang\textsuperscript{\dag},
    Binwu Wang\textsuperscript{*,\dag},
    Pengkun Wang\textsuperscript{\dag},
    Yang Wang\textsuperscript{*,\dag}}
\IEEEauthorblockA{\textsuperscript{\dag}~\textit{University of Science and Technology of China (USTC)}, Hefei, China\\
Email: \{JiamingMa, wgj, shenghuang, yangkuo\}@mail.ustc.edu.cn, \{wbw2024, pengkun, angyan\}@ustc.edu.cn}
\thanks{Dr. Binwu Wang and Prof. Yang Wang are the corresponding authors.}

}

\maketitle

\begin{abstract}
Due to the profound impact of air pollution on human health, livelihoods, and economic development, air quality forecasting is of paramount significance.  Initially, we employ the causal graph method to scrutinize the constraints of existing research in comprehensively modeling the causal relationships between the air quality index (AQI) and meteorological features.  In order to enhance prediction accuracy, we introduce a novel air quality forecasting model, AirCade, which incorporates a causal decoupling approach.  AirCade leverages a spatiotemporal module in conjunction with knowledge embedding techniques to capture the internal dynamics of AQI.  Subsequently, a causal decoupling module is proposed to disentangle synchronous causality from past AQI and meteorological features, followed by the dissemination of acquired knowledge to future time steps to enhance performance.  Additionally, we introduce a causal intervention mechanism to explicitly represent the uncertainty of future meteorological features, thereby bolstering the model's robustness.  Our evaluation of AirCade on an open-source air quality dataset demonstrates over 20\% relative improvement over state-of-the-art models. Our source code is available at https://github.com/PoorOtterBob/AirCade.
\end{abstract}

\begin{IEEEkeywords}
air quality forecasting, spatiotemporal forecasting, deep learning, urban computing
\end{IEEEkeywords}

\section{Introduction}

Air quality plays a crucial role in ensuring human health and well-being. According to the World Health Organization (WHO), air pollution is one of the leading causes of death globally, resulting in 7 million deaths annually \cite{vallero2014fundamentals}. The challenge of air quality prediction is garnering increasing attention from both academia and industry. 

Over the past decades, deep learning-based models for air quality prediction, in particular, have achieved remarkable results. In particular, with the rise of GCN across various domains~\cite{wan2024federated,zou2025loha,wan2024epidemiology,huang2023crossgnn,yangimproving,chen2025stcontext}, researchers have begun using spatiotemporal graph convolutional networks for spatiotemporal data prediction~\cite{wang2023pattern,wang2023knowledge,miao2024unified,chen2024expand,liu2024dynamic,zhang2024modeling,wang2024stone,wang2024towards,zhang2024meta}. These methods utilize historical observations and auxiliary covariates (such as weather information) to extract high-dimensional spatiotemporal representations of atmospheric conditions. For example, PM2.5-GCN~\cite{wang2020pm2} introduced a basic GCN to model dependencies between stations and combined it with LSTM to capture temporal dynamics in long sequences. To improve prediction accuracy, researchers have also introduced attention mechanisms. For example, AirFormer~\cite{liang2023airformer} designed a two-stage self-attention mechanism to simultaneously learn spatiotemporal dynamics and inherent uncertainties in atmospheric data.

\begin{figure}[!t]
	\centering
	\includegraphics[width=0.45\textwidth]{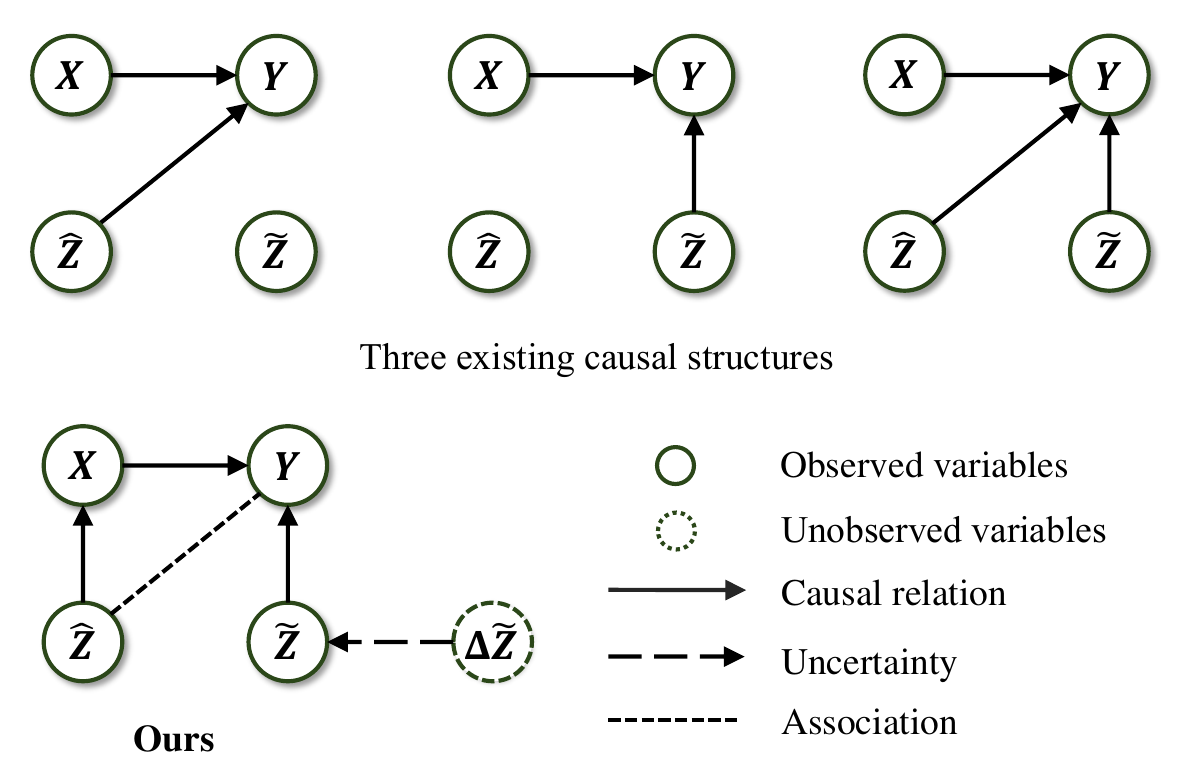}
	\caption{Three different causal structures with our proposed perspective. $\mathbf{X}$ and $\mathbf{Y}$ represent the past and future AQI respectively. $\mathbf{\hat{Z}}$ and $\mathbf{\widetilde{Z}}$  represent the past and future weather conditions respectively.} 
	\label{scm}\vspace{-0.15in}
\end{figure}
Accurate air quality prediction relies on modeling the causal relationships between AQI and weather conditions. Existing models can be categorized into three causal structures, as shown in Fig.\ref{scm}. Conventional spatio-temporal graph convolutional networks ~\cite{shao2022decoupled, jiang2023pdformer} simply concatenate air quality data as input feature channels. While some models use shallow neural networks (like fully connected layers) to encode future weather information during the decoding phase and concatenate it with high-dimensional representations of air quality data to generate predictions, these approaches fail to explicitly model the synergistic causal relationships between AQI and weather conditions.

In this study, we introduce a novel \underline{\textbf{Ca}}usal \underline{\textbf{De}}coupling  \underline{\textbf{Air}} Quality Forecasting model (AirCade), which explicitly models the complex causal relationships between atmospheric quality data and weather information. AirCade integrates innovative temporal and spatial Transformers and employs spatiotemporal embedding techniques to capture the inherent dynamics of AQI. Each Transformer features an air causal decoupling module, including a causal diffusion module that models the synchronous causal relationships between past AQI and meteorological features across both temporal and spatial dimensions. The knowledge gained is propagated to future time steps for accurate predictions. Furthermore, we incorporate a causal intervention mechanism to perturb this propagation, allowing uncertainty to be integrated into the model and enhancing its robustness. Experimental results on real world data show that our model achieves competitive performance. Our contributions are summarized as follows:

\begin{itemize}
    \item We introduce a novel causal structure that covers a comprehensive modeling of the causal relationships between AQI and meteorological features. Leveraging this framework, we develop a causal decoupling air forecasting model called AirCade.
    \item AirCade includes a causal decoupling module to model the causal relationships between different variables. We also design a causal intervention  mechanism to enable the model learn more robust causal relationships.
    \item  Through experiments conducted on the currently largest open-access air quality dataset, our model is demonstrated to achieve competitive performance.
\end{itemize}

\section{Problem definition}

We use $\mathrm{x}_t\in\mathbb{R}^{N\times c}$ to denote the air quality index collected from  from $N$ stations at time-step $t$, where $c$ is the number of features such as PM$_{2.5}$ or NO$_{2}$. meteorological features $\mathrm{z}_t\in\mathbb{R}^{N\times f}$ is collected from all stations where $f$ is the number of features such as weather and wind speed. 

Given AQI in past $T$ time steps $\mathbf{X}=\{\mathrm{x}_{1},...,\mathrm{x}_{T}\}\in\mathbb{R}^{T\times N\times c}$, past meteorological features $\mathbf{\hat{Z}}=\{\mathrm{z}_{1},...,\mathrm{z}_{T}\}$, and further meteorological features $\mathbf{\widetilde{Z}}=\{\mathrm{z}_{T+1},...,\mathrm{z}_{T+T_P}\}$~\footnote{We simulate future weather forecasts by adding noise.}, air quality forecasting aims to predict further AQI next $T_P$ time steps $\mathbf{Y}=\{\mathrm{x}_{T+1},...,\mathrm{x}_{T+T_P}\}\in\mathbb{R}^{T_{P}\times N\times c}$. 
\section{Methodology}
\begin{figure}
    \centering
    \includegraphics[width=0.45\textwidth]{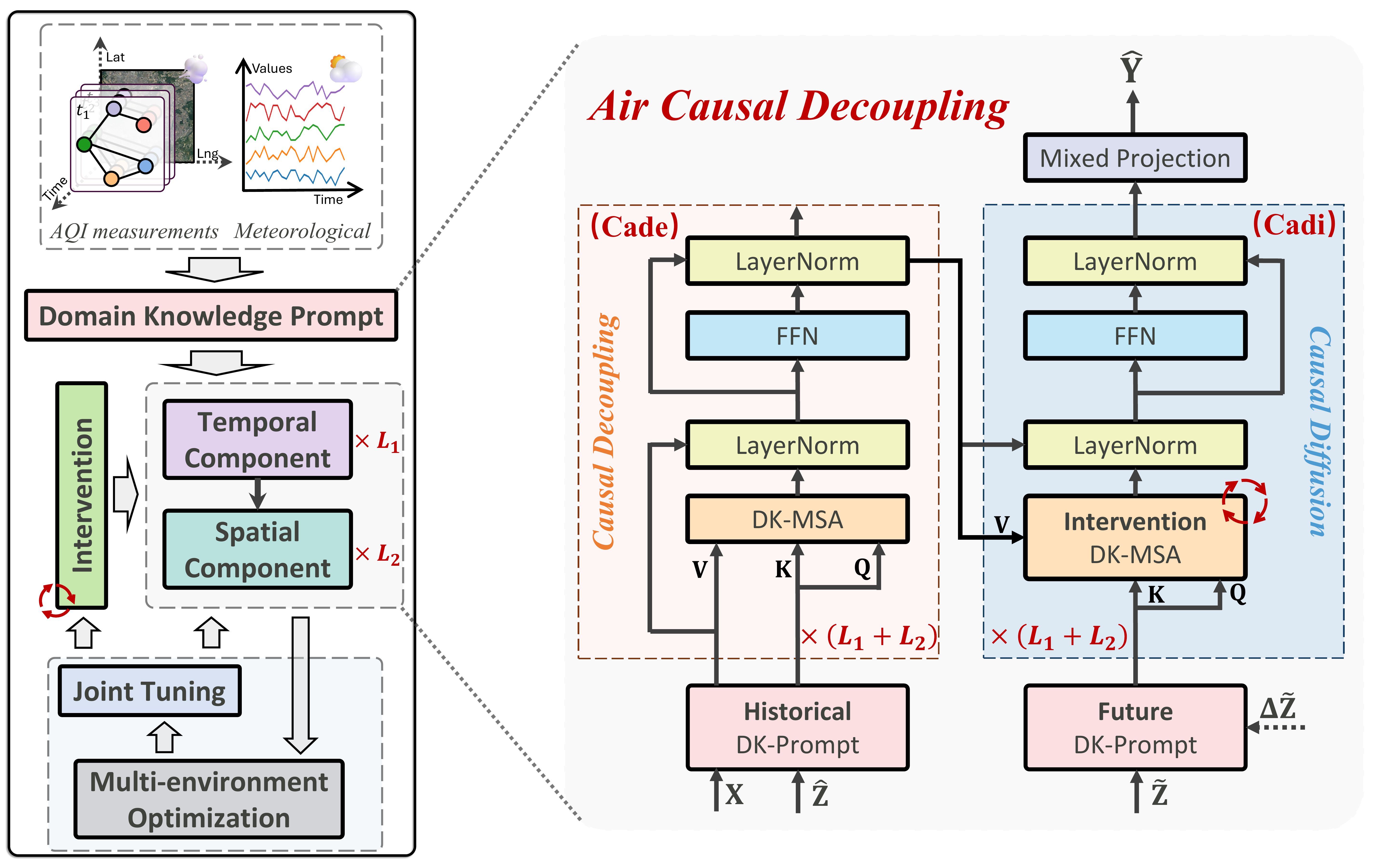} 
    \caption{Details of the proposed model (AirCade), which consists of spatial and temporal transformers. Each Transformer layer contains a causal decoupling module that explicitly models the relationship between AQI and weather. }
    \label{overmodel} \vspace{-0.15in}
\end{figure}

The details of Aircade are shown in Fig.\ref{overmodel}. Our model consists of three parts: a spatiotemporal learning module with domain knowledge prompt technology, an air causal decoupling transformer, and the causal intervention mechanism. 

\subsection{Spatiotemporal Learning Module}
In order to learn spatiotemporal dynamics of AQI, we employ the Domain Knowledge Prompt (DK-Prompt) technology from computer vision~\cite{jia2022visual} and natural language processing~\cite{vaswani2017attention} by encoding prior information using embedding techniques~\cite{zhou2021informer}. For past and future horizons, we use two different sets of embeddings $\bm{e}_\text{D}$ 
and  $\tilde{\bm{e}}_\text{D}\in\mathbb{R}^{N_{T}\times d_P}$ to encode dynamic information of each time step, where $N_T$ means daily sampling frequency. In the spatial dimension, we leverage adaptive station-level embeddings $\bm{e}_\text{S}$ and  $\tilde{\bm{e}}_\text{S}\in\mathbb{R}^{N\times d_P}$ to learn air quality patterns of different stations. We also use position embedding $\bm{e}_\text{P}$ and $\tilde{\bm{e}}_\text{P}\in \mathbb{R}^{T\times N\times d_s}$ to encode sequential information of input. Then we integer embedded information into three variables $\mathbf{X}$, $\mathbf{\hat{Z}}$, and $\mathbf{\widetilde{Z}}$.  Taking $\mathbf{X}$ as an example, we first use a MLP layer to capture internal spatiotemporal dynamics and concatenate the output with the embedding vectors:
\begin{align}
    \mathbf{H}_{v} = [\operatorname{MLP}_1(\mathbf{X})+\bm{e}_P\|\bm{e}_D\| \bm{e}_S]\in\mathbb{R}^{T\times N\times d_{m}},
\end{align}
where $\mathbf{H}_{v}$ means AQI representation. Similarly, we can get past and future meteorological representation:  $\mathbf{H}_{z}$ and $\tilde{\mathbf{H}}_{z}$.

\subsection{Air Causal Decoupling Module}
This module consists of $L_1$ temporal components followed by $L_2$ spatial transformer components. Each component consists of two parts: the \textbf{causal decoupling} layer (Cade) as encoder and the \textbf{causal diffusion} layer (Cadi) as decoder, where Domain knowledge multi-head self-attention
mechanism (DK-MSA) is used as their attention function.

\subsubsection{DK-MSA} 
\begin{figure}[!t]
	\centering
	\includegraphics[width=0.43\textwidth]{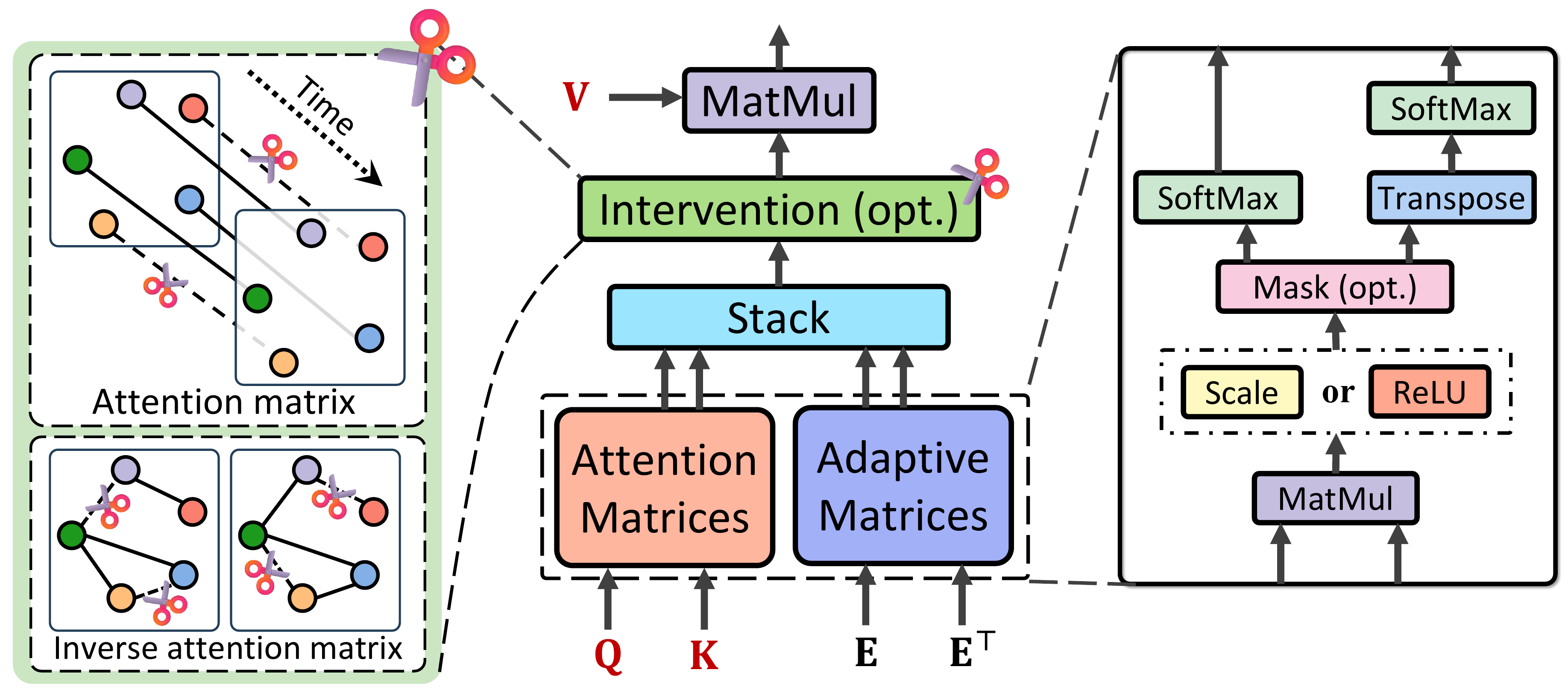}
	\caption{Domain knowledge multi-head self-attention mechanism (DK-MSA). Adaptive adjacency matrices coefficients and gated mechanism are incorporated into the self-attention function, alongside masking of coefficients to simulate environmental intervention.}
	\label{com111}
\end{figure}

Domain knowledge multi-head self-attention mechanism (DK-MSA) is used to model the causal relationship between meteorological features and AQI. As shown in Fig.\ref{com111}, given Query vector $\mathbf{Q}\in\mathbb{R}^{N\times d}$, Key vector $\mathbf{K}\in\mathbb{R}^{N\times d}$,  and Value vector $\mathbf{V}\in\mathbb{R}^{N\times d}$, it 
 $\mathcal{H}\left(\cdot\right)$ adopts the diffusion attention mechanism, which calculates attention scores from two paths and integrates an adaptive attention learning strategy: 
\begin{align}
\mathcal{H}&\left(\mathbf{Q},\mathbf{K},\mathbf{V}\right)=\left[\mathbf{A}_{1},\mathbf{A}_{2},\mathbf{A}_{3},\mathbf{A}_{4}\right]\mathbf{V},\\
    \mathbf{A}_{1}&=\mathcal{S}\left(\alpha \mathbf{Q}\mathbf{K}^\top\right),~~\mathbf{A}_{2}=\mathcal{S}\left(\operatorname{ReLU}\left(\mathbf{E}_{1}\mathbf{E}_{2}^\top\right)\right),\\
    \mathbf{A}_{3}&=\mathcal{S}\left(\alpha \mathbf{K}\mathbf{Q}^\top\right)^{\top},\mathbf{A}_{4}=\mathcal{S}\left(\operatorname{ReLU}\left(\mathbf{E}_{2}\mathbf{E}_{1}^\top\right)\right)^\top,
\end{align}
where $\mathbf{E}_{2} \in\mathbb{R}^{N\times d_e}$ and $\mathbf{E}_{1} \in\mathbb{R}^{N\times d_e}$ are learnable parameters. $\mathbf{A}_{i}  \in\mathbb{R}^{N\times N}, i\in\{1, 2, 
 3, 4\}$ means different attention coefficients. $\alpha=\sqrt{d_{m}}$ is a scaling factor, and $\mathcal{S}\left(\cdot\right)$ is a $\operatorname{softmax}$ operation. We use a multi-head attention mechanism with $K_h$ heads to enhance the representation ability:
\begin{align}
    \operatorname{DK-MSA}&\left(\mathbf{Q},\mathbf{K},\mathbf{V}\right)=\left[\mathcal{H}_{1}\|...\|\mathcal{H}_{K_h}\right]W
\end{align}
where $W$ is a learnable parameter. In the temporal transformer layer, we design a gate to filter out redundant information,  and $\mathcal{H}_{i}$ can be computed:
\begin{align}
\mathcal{H}_{i}=\mathcal{T}\left(\mathcal{H}\left(\mathbf{Q},\mathbf{K},\mathbf{V}\right)W_{i}^{t}\right)\odot\sigma\left(\mathcal{H}\left(\mathbf{Q},\mathbf{K},\mathbf{V}\right)W_{i}^{s}\right),
\end{align}
where $\mathcal{T}\left(\cdot\right)$ is the $\operatorname{tanh}$ function and $\sigma\left(\cdot\right)$ is the $\operatorname{Sigmoid}$ function with learnable parameters $W_{i}^{t}$ and $W_{i}^{s}\in\mathbb{R}^{d_{m}\times d_{m}}$. In the spatial transformer layer, $\mathcal{H}_{j}$ is the linear version
\begin{align}
\mathcal{H}_{j}=\mathcal{H}\left(\mathbf{Q},\mathbf{K},\mathbf{V}\right)W_{j},
\end{align}
 where $W_{j}\in\mathbb{R}^{d_{m}\times d_{m}}$ is a learnable parameter.

\subsubsection{Causal decoupling}  the Cade layer is used to model the causal relationship between meteorological features and AQI that is aligned in time steps. Specifically, it taks historical meteorological representation $\mathbf{H}_{z}$ as $\mathbf{Q}$ and $\mathbf{K}$ vectors, and $\mathbf{V}$ vector is the output of the previous Cade layer, which is denoted as $\mathbf{V}_{l-1}$ and $\mathbf{V}_{0}=\mathbf{H}_{v}$ where $\mathbf{H}_{v}$ is the high-dimensional mapping of AQI data. These vectors are taken input into DK-MSA to generate the output $\mathbf{O}_l$:
\begin{align}    &\mathbf{O}_l=\operatorname{LN}\left(\operatorname{MLP}\left(\mathbf{V}_{l}\right)+\mathbf{V}_{l}\right),\\
    &\mathbf{V}_{l}=\operatorname{LN}\left(\operatorname{DK-MSA}\left(\mathbf{H}_{z}, \mathbf{H}_{z}, \mathbf{V}_{l-1} \right) + \mathbf{V}_{l-1}  \right).
\end{align}
where $\operatorname{LN}\left(\cdot\right)$ is layer normalization operation.

\subsubsection{Causal diffusion} this layer diffusion future meteorological features for generating predictions. Specifically, the output of $l$-th Cade layer $\mathbf{O}_l$ is used as 
Value vector $\tilde{\mathbf{V}}_l$ of this layer. And $\mathbf{Q}$ and $\mathbf{K}$ vectors are future meteorological representation $\tilde{\mathbf{H}}_{z}$. The output of this layer can be computed: 
\begin{align}    &\tilde{\mathbf{O}}_l=\operatorname{LN}\left(\operatorname{MLP}\left(\tilde{\mathbf{V}}_{l}\right)+\tilde{\mathbf{V}}_{l}\right),\\
&\tilde{\mathbf{V}}_{l}=\operatorname{LN}\left(\operatorname{DK-MSA}\left(\tilde{\mathbf{H}}_{z}, \tilde{\mathbf{H}}_{z}, \mathbf{O}_l \right) + \mathbf{O}_l  \right).
\end{align}

\subsubsection{Predictor} After the learning of air causal decoupling module, we use the fully connected layer to encode $\tilde{\mathbf{O}}_{L_2}$ for generating predictions with learnable parameters $W_o\in\mathbb{R}^{d_{m}\times c}$ and $b_o\in\mathbb{R}^{c}$:
\begin{align}
\hat{\mathbf{Y}}=\tilde{\mathbf{O}}_{L_2}W_o+b_o\in\mathbb{R}^{T_{P}\times N\times c},
\end{align}

\subsection{Causal Intervention Mechanism}
Future meteorological data is inherently uncertain due to inaccuracies in weather forecasting systems or unpredictable factors. We consider incorporating this uncertainty into the analysis to achieve robust predictions. We propose a causal intervention mechanism. Specifically, we employ $K$ sets of binary masking operators $\mathbb{M}=\left\{\mathbf{M}_{1},...,\mathbf{M}_{K}\right\}$, each set containing two kinds of masking matrices $\mathrm{M}_k^1\in\left[0,1\right]^{T_{P}\times T_{P}}$ and inverse attention $\mathrm{M}_k^2\in\left[0,1\right]^{N\times N}$. $\mathrm{M}_k^1$ is applied to every DK-MSA layer in the temporal component, and its shape is consistent with the calculated attention coefficient shape. We get the intervention attention by dotting $\mathrm{M}_k^1$ with the four attention coefficients for output, as shown in Fig.~\ref{com111}. Similarly, $\mathrm{M}_k^2$ is applied to the spatial component. To improve the robustness of the model, we maximize the variance of the mask to cover as many differences as possible.
\begin{equation}
\begin{gathered}
\min _{\Theta} \operatorname{Var}\left\{\mathcal{L}\left(\mathbf{Y}\mid \mathbb{M}^*, \Theta\right)\right\}+\beta \mathcal{L}\left(\mathbf{Y} \mid  \mathbb{M}^*, \Theta\right),\\
\text{s. t.} \quad \mathbb{M}^*=\underset{k \in \{1, \cdots, K\}}{\operatorname{argmax}} \operatorname{Var}\left\{\mathcal{L}\left(\mathbf{Y}\mid \mathbf{M}_{k}, \Theta\right)\right\},
\end{gathered}
\end{equation}
where Var($\cdot$) means the loss variance, $\beta$ is a trade-off to balance two loss terms.
\section{Experiment}
\subsection{Experiment Setting}
\subsubsection{Datasets}
We utilize two subsets from the fully open-sourced air quality dataset KnowAir \cite{wang2020pm2}. This dataset captures PM$_{2.5}$ feature from 184 
cities in China along with their corresponding 13 meteorological attributes. For the uncertainty of future meteorological features, we introduce standard normally distributed noise $\Delta\tilde{\mathbf{Z}}\sim\mathcal{N}\left(
0,1\right)$ to $\tilde{\mathbf{Z}}$.

\subsubsection{Setting} The length of the input and prediction windows are both set to 24. The optimizer utilizes RMSprop~\cite{tieleman2017divide}, configured with a learning rate $5\times10^{-4}$. $L_1$ and $L_2$ are both equal to $3$. The head number $h=8$ and the environment number $K=3$. We used a variety of metrics for evaluation, including MAE, RMSE, MAPE, which are commonly used in spatiotemporal data prediction~\cite{wang2024condition,yiget}, and some specific ones to atmospheric prediction. We compare spatiotemporal graph prediction models with specialized air quality prediction models. Spatiotemporal graph prediction models include STID~\cite{shao2022spatial}, STGCN~\cite{yu2017spatio}, GWNet~\cite{wu2019graph}, CaST~\cite{xia2024deciphering}, STAEformer~\cite{liu2023spatio}, and STTN~\cite{xu2020spatial}. The later models include GC-LSTM~\cite{qi2019hybrid},  PM$_{2.5}$GNN~\cite{wang2020pm2}, GAGNN~\cite{chen2023group}, DeepAir~\cite{yi2018deep}, Airformer~\cite{liang2023airformer}, and MGSFformer~\cite{yu2025mgsfformer}. 

\begin{table*}[!htbp]
  \belowrulesep=1.5pt
  \aboverulesep=0.5pt
  \setlength{\tabcolsep}{2.5pt}
  \centering
  \caption{Experiments on datasets. We bold the best-performing model results in {\color{red}\textbf{red}} and underline the suboptimal model results in \textcolor{blue}{\underline{blue}}. }
    \resizebox{\linewidth}{!}{
    \begin{tabular}{ccccccc|cccccc}
    \toprule
    \multirow{2}[1]{*}{Method} & \multicolumn{6}{c}{KnowAir (2015)}            & \multicolumn{6}{c}{KnowAir (2017)} \\
\cmidrule{2-13}          & MAE $\downarrow$   & RMSE $\downarrow$  & MAPE (\%) $\downarrow$  & CSI (\%) $\uparrow$   & POD (\%) $\uparrow$   & FAR (\%) $\downarrow$   & MAE $\downarrow$   & RMSE $\downarrow$  & MAPE (\%) $\downarrow$  & CSI (\%) $\uparrow$   & POD (\%) $\uparrow$   & FAR (\%) $\downarrow$ \\
    \midrule
    \midrule
    STID~\cite{shao2022spatial}& 18.22±0.89 & 30.91±1.06 & 41.58±3.22 & 69.58±3.13 & 84.98±2.09 & 20.66±2.35 & 14.75±0.67 & 24.98±0.89 & 44.06±1.97 & 60.59±3.57 & 73.12±2.09 & 22.04±1.82 \\
    STGCN~\cite{yu2017spatio}& 17.99±0.35 & 30.44±0.89 & 42.84±3.08 & 70.42±1.42 & 87.98±1.30 & 22.08±1.94 & 14.40±0.37 & 24.50±0.48 & 42.57±1.63 & 60.96±4.24 & 71.65±4.13 & 19.66±2.38 \\
    GWNet~\cite{wu2019graph}& 17.91±0.41 & 30.45±0.97 & 40.58±2.12 & 69.82±1.89 & 85.47±1.70 & 20.77±1.40 & 14.40±0.32 & 24.59±0.44 & 42.23±1.14 & 60.67±2.61 & 71.47±3.07 & 19.94±1.89 \\
    CaST~\cite{xia2024deciphering}& 18.54±0.88 & 31.18±1.84 & 45.10±3.76 & 69.56±3.68 & 85.38±1.83 & 21.03±1.96 & 15.40±0.53 & 25.77±0.68 & 48.82±3.94 & 58.57±1.60 & 70.96±1.87 & 22.96±2.57 \\
    STAEformer~\cite{liu2023spatio}& 18.78±0.76 & 31.51±1.23 & 46.40±3.56 & 69.59±2.74 & \textcolor{blue}{\underline{89.27±1.55}} & 24.06±1.53 & 14.59±0.60 & 24.79±0.61 & 42.38±3.14 & 60.33±2.17 & 72.25±3.60 & 21.47±2.34 \\
    STTN~\cite{xu2020spatial}& 21.28±0.75 & 35.41±1.72 & 52.61±2.89 & 64.09±3.09 & 81.93±2.96 & 25.36±1.57 & 16.03±0.51 & 26.99±0.66 & 46.27±3.86 & 54.14±4.10 & 64.74±4.66 & 23.22±2.14 \\
    \midrule
    GC-LSTM~\cite{qi2019hybrid}& 20.13±0.41 & 34.23±1.35 & 44.76±2.49 & 66.56±1.64 & 86.61±1.73 & 25.81±1.08 & 16.64±0.43 & 27.96±0.42 & 46.79±1.54 & 55.02±2.21 & 70.84±2.98 & 28.87±1.71 \\
    PM2.5-GNN~\cite{wang2020pm2}& 19.18±0.50 & 32.91±1.50 & 38.69±2.42 & 69.99±2.84 & 87.69±1.30 & 22.38±1.44 & 14.93±0.42 & 25.42±0.90 & 40.66±2.98 & 58.18±2.87 & 69.12±3.01 & 21.39±2.59 \\
    GAGNN~\cite{chen2023group}& 18.67±0.69 & 31.39±1.14 & 45.72±3.89 & 69.10±2.43 & 85.76±2.24 & 21.94±1.31 & 15.09±0.55 & 25.50±0.90 & 44.57±2.44 & 57.48±2.10 & 66.14±4.90 & 18.56±1.46 \\
    DeepAir~\cite{yi2018deep}& \textcolor{blue}{\underline{17.44±0.31}} & \textcolor{blue}{\underline{30.30±0.97}} & \textcolor{blue}{\underline{32.88±1.40}} & \textcolor{blue}{\underline{71.87±2.57}} & 88.40±1.19 & \textcolor{blue}{\underline{20.20±2.28}} & \textcolor{blue}{\underline{13.57±0.39}} & \textcolor{blue}{\underline{23.22±0.58}} & \textcolor{blue}{\underline{38.28±1.81}} & \textcolor{blue}{\underline{64.03±2.49}} & \textcolor{blue}{\underline{74.70±1.93}} & \textcolor{blue}{\underline{18.25±1.35}} \\
    Airformer~\cite{liang2023airformer}& 17.99±0.42 & 31.12±1.11 & 33.43±1.45 & 70.54±2.12 & 87.10±1.08 & 21.12±1.88 & 14.22±0.45 & 24.01±0.62 & 40.50±1.67 & 63.12±1.45 & 74.17±1.24 & 18.65±1.42 \\
    MGSFformer~\cite{yu2025mgsfformer}& 18.24±0.43& 31.17±1.50 & 35.48±1.54 & 70.88±2.24 & 86.53±0.93 & 20.86±2.48 & 13.89±0.28 & 23.97±0.43 & 39.45±2.18 & 62.73±1.98 & 74.00±1.72 & 19.24±1.10 \\
    \midrule
    \textbf{Ours} & \textcolor{red}{\textbf{14.60±0.36}} & \textcolor{red}{\textbf{26.22±0.86}} & \textcolor{red}{\textbf{24.59±2.12}} & \textcolor{red}{\textbf{77.27±2.06}} & \textcolor{red}{\textbf{90.44±1.36}} & \textcolor{red}{\textbf{13.23±1.54}} & \textcolor{red}{\textbf{11.29±0.24}} & \textcolor{red}{\textbf{20.09±0.46}} & \textcolor{red}{\textbf{28.33±2.33}} & \textcolor{red}{\textbf{72.41±2.19}} & \textcolor{red}{\textbf{80.73±2.37}} & \textcolor{red}{\textbf{14.37±1.68}} \\
    \bottomrule
    \end{tabular}}%
  \label{mainexp}%
\end{table*}%

\subsection{Main Comparison}
As shown in Table \ref{mainexp}, in the spatiotemporal prediction family, GWNet and STGCN models exhibit outstanding performance, because they utilize advanced spatiotemporal graph convolutional networks to effectively model the causal relationships between historical and predicted AQI. GAGNN introduces differentiable partible networks to discover potential dependencies. Airformer and MGSFformer utilize Transformers to model the associations between regions globally. Surprisingly, DeepAir achieves lower predictive performance as it integrates future atmospheric information, which significantly impacts further air quality. Attributed to our comprehensive modeling of the causal relationships between different variables, our AirCade demonstrates superior performance. 


\subsection{Ablation Study}
We design seven variants for the ablation study to explore the effectiveness of each component in AirCade. The variant "w/o prompt" removes the knowledge prompt technology. "w/o adp", "w/o agg", and "w/o diff" eliminate adaptive, aggregating and diffusion attention in DK-MSA respectively. "w/o cade" uses only the cadi module from future meteorological features, omitting Cade. "w/o es" means that we use the value output of last layer of Cade to be the final prediction, and "w/o intv" removes the proposed causal intervention mechanism.

\begin{figure}[!h]
	\centering
	\includegraphics[width=0.45\textwidth]{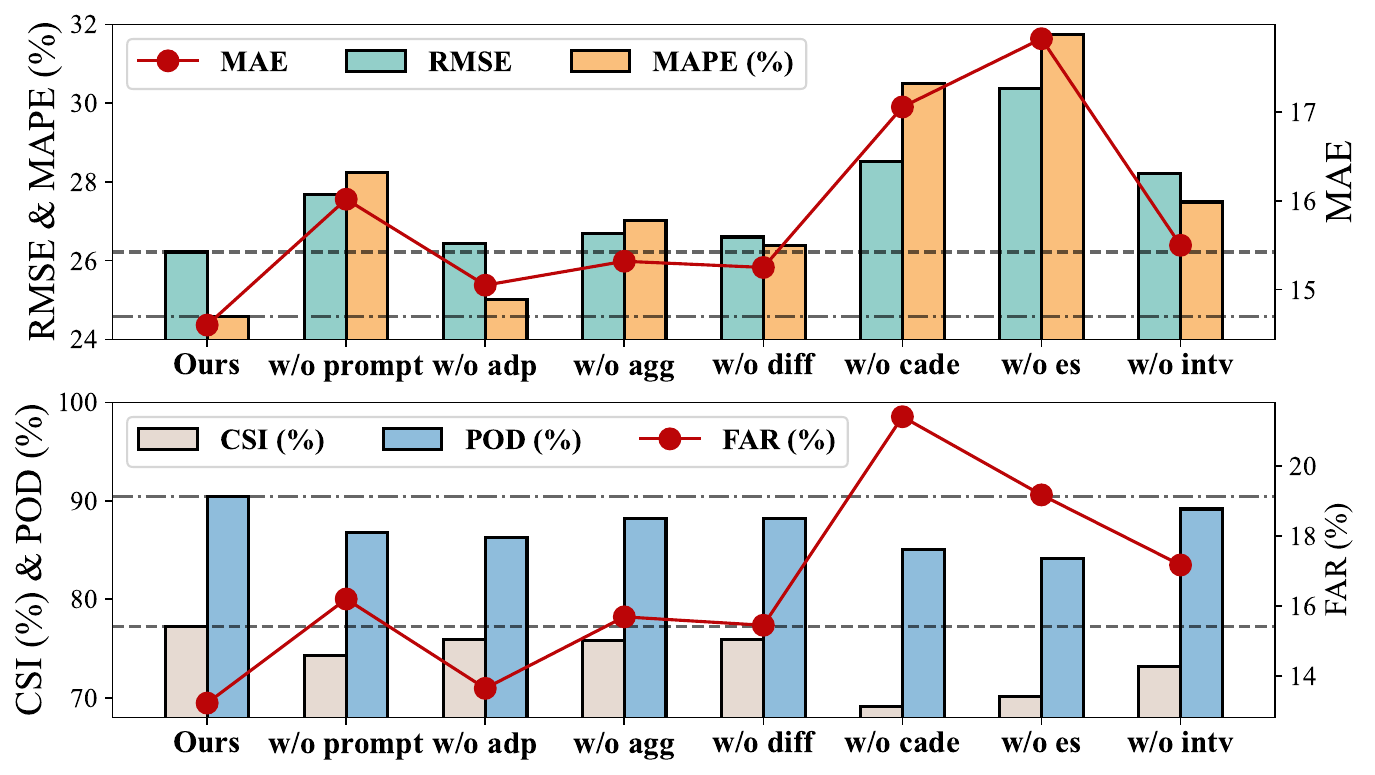}
	\caption{Ablation results on KnowAir dataset.}
\label{ablation}
\end{figure}

As shown in Fig.\ref{ablation}, "w/o cade" exhibits lower predictive performance, indicating that integrating various prior knowledge can enhance prediction accuracy. "w/o es" performs the worst, demonstrating the necessity of incorporating future meteorological information. In conclusion, each component of the model is effective.
\begin{figure}[!h]
	\centering
	\includegraphics[width=0.48\textwidth]{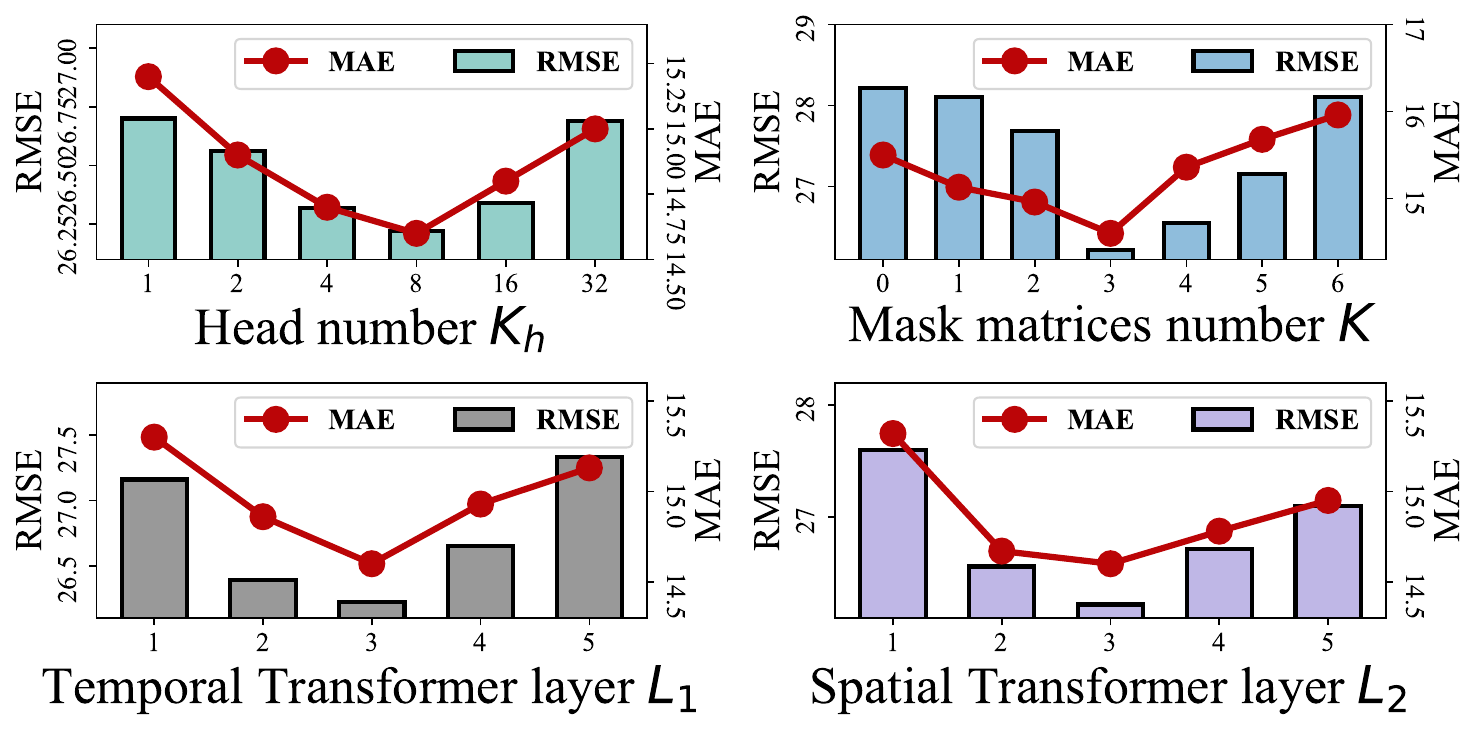}
	\caption{Hyperparameter sensitivity results on KnowAir dataset.}\vspace{-0.1in}
\label{hyper}
\end{figure}
\subsection{Hyperparameter Sensitivity Analysis }
We analyze the impact of four key hyperparameters on KnowAir dataset. The results are shown in Fig.\ref{hyper}. when $K_h$ is equal to 8  and Mask matrix number $K_e$ is set to 3, AirCade gets the best performance. For the number of Transformer layer $L_1$ and $L_2$,  they are both set to 3. When this value is smaller, small-scale parameters limit the representational power of the model, while too many layers increase complexity and training difficulty, hindering practical application.

\subsection{Prompt Embedding Visualizations}
\begin{figure}[!h]
	\centering
	\includegraphics[width=0.45\textwidth]{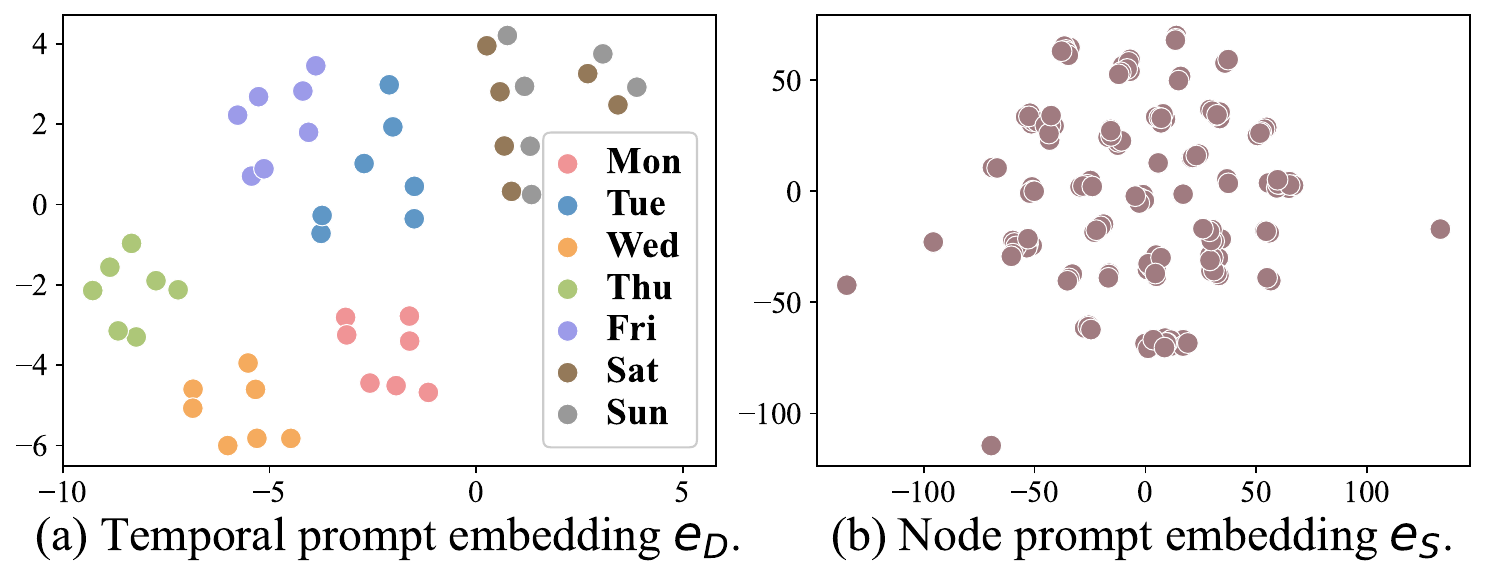}
	\caption{Prompt embedding visualizations.}
\label{cases} 
\end{figure}

We extract trained prompt embedding vectors from AirCade for visualization. As shown in Fig.\ref{cases}, We can see that the temporal embeddings $e_D$ effectively distinguish the air features of different days within a week, thereby modeling temporal dynamics effectively. On the other hand, within spatial embeddings, the features of the nodes cluster and distribute well, with clear boundaries. The air quality characteristics among nodes within these clusters are likely to be similar.

\section{conclusion}
We develop an air quality forecasting model, AirCade, designed to comprehensively model the causal relationships among air features. This model integrates a spatio-temporal prompt layer for dynamic learning, a causal decoupling module to capture causal associations, and a intervention mechanism to enhance learning robustness. Experimental results have demonstrated the effectiveness of the AirCade.

\section*{ACKNOWLEDGEMENTS}
This paper is partially supported by the National Natural Science
Foundation of China (No.12227901, No.62072427), the Project of Stable Support for Youth Team in Basic Research Field, CAS (No.YSBR005), Academic Leaders Cultivation Program, USTC.
\clearpage
\bibliographystyle{ieeetr}
\bibliography{icassp}

\end{document}